\RequirePackage{etex} %% **************** add this for ARXIV submission
\documentclass[runningheads,twocolumn,a4paper,10pt]{llncs}

\usepackage[maxnames=6,firstinits=true,doi=false,url=true,isbn=false]{biblatex}
\newcommand{\pathToOtherFiles}{other}

\usepackage{etex}
\reserveinserts{28}
\pdfoutput=1
\usepackage[a4paper, total={7in, 10in}]{geometry}
\usepackage[linesnumbered,lined,boxed,commentsnumbered,ruled]{algorithm2e}

\SetAlFnt{\small}
\SetAlCapFnt{\small}
\SetAlCapNameFnt{\small}
\SetAlCapHSkip{0pt}
\IncMargin{-\parindent}
\usepackage{ textcomp }
\usepackage{ comment }
\usepackage[color=yellow!60,textsize=footnotesize,obeyDraft,draft]{todonotes}
\usepackage{colortbl}
\usepackage{booktabs}
\usepackage{capt-of}
\usepackage{makecell}
\usepackage{multirow}
\usepackage{listings}
\usepackage{graphicx}
\usepackage{amsmath}
\usepackage[font=footnotesize,labelfont=bf]{subcaption}
\usepackage[capitalize]{cleveref}
\usepackage{tikz}
\usepackage{pgfplots}
\usepackage[para,online,flushleft]{threeparttable}
\usepackage{wasysym}
\usepackage{amssymb}
\usepackage{enumitem}
\usetikzlibrary{arrows}
\usepackage{breakcites}
\usepackage{array}
\usepackage{color}
\usepackage{balance} %balance columns
\usepackage{lastpage}
\usepackage{dblfloatfix}
\usepackage{authblk}
\usepackage{siunitx}
\usepackage{xurl} %for long URLs

\usepackage[hyphenbreaks]{breakurl}

\begin{document} 
% \mainmatter

\title{On Determinism of Game Engines used for Simulation-based Autonomous Vehicle Verification}

\titlerunning{On Determinism of Game Engines... }

\author{Greg Chance\inst{1,3}, Abanoub Ghobrial\inst{1,3}, Kevin McAreavey\inst{1,3}, S\'everin Lemaignan\inst{2,3}, Tony Pipe\inst{2,3}, Kerstin Eder\inst{1,3}}

\authorrunning{G. Chance, A. Ghobrial et al.}
\institute{University of Bristol, Bristol, UK \and University of the West of England, Bristol, UK \and Bristol Robotics Laboratory, Bristol, UK}
\tocauthor{Authors' Instructions}

\maketitle
\let\thefootnote\relax\footnotetext{\textit{Greg Chance and Abanoub Ghobrial contributed equally to this paper, Greg Chance is the corresponding author.}
Greg Chance (e-mail: greg.chance@bristol.ac.uk), 
Abanoub Ghobrial (e-mail: abanoub.ghobrial@bristol.ac.uk), 
Kevin McAreavey (e-mail: kevin.mcareavey@bristol.ac.uk), 
and 
Kerstin Eder (e-mail: kerstin.eder@bristol.ac.uk) 
are with the Trustworthy Systems Lab, Department of Computer Science, University of Bristol, Merchant Ventures Building, Woodland Road,  Bristol, BS8 1UQ, United Kingdom. 
S\'everin Lemaignan (e-mail: severin.lemaignan@brl.ac.uk)
and
Tony Pipe (e-mail: tony.pipe@brl.ac.uk), 
are with the Bristol Robotics Lab, T Block, University of the West of England, Frenchay, Coldharbour Ln, Bristol, BS34 8QZ, United Kingdom.}
\makeatletter
\renewcommand\subsubsection{\@startsection{subsubsection}{3}{\z@}%
                       {-18\p@ \@plus -4\p@ \@minus -4\p@}%
                       {4\p@ \@plus 2\p@ \@minus 2\p@}%
                       {\normalfont\normalsize\bfseries\boldmath
                        \rightskip=\z@ \@plus 8em\pretolerance=10000 }}
\makeatother

% \begin{multicols}{2}

\textbf{\textit{Abstract}Game engines are increasingly used as simulation platforms by the autonomous vehicle community to develop vehicle control systems and test environments. A key requirement for simulation-based development and verification is determinism, since a deterministic process will always produce the same output given the same initial conditions and event history. Thus, in a deterministic simulation environment, tests are rendered repeatable and yield simulation results that are trustworthy and straightforward to debug. However, game engines are seldom deterministic. This paper reviews and identifies the potential causes and effects of non-deterministic behaviours in game engines. A case study using CARLA, an open-source autonomous driving simulation environment powered by Unreal Engine, is presented to highlight its inherent shortcomings in providing sufficient precision in experimental results. Different configurations and utilisations of the software and hardware are explored to determine an operational domain where the simulation precision is sufficiently high i.e.\ variance between repeated executions becomes negligible for development and testing work. Finally, a method of a general nature is proposed, that can be used to find the domains of permissible variance in game engine simulations for any given system configuration.
}

% ***************************************************
%  Main Body
% ***************************************************
% ***************************************************
%  INTRODUCTION
% ***************************************************
\section{Introduction} \label{s:introduction}
Simulation-based verification of autonomous driving functionality is a promising counterpart to costly on-road testing, that benefits from complete control over (virtual) actors and their environment.
Simulated tests aim to provide evidence to developers and regulators of the functional safety of the vehicle or its compliance with commonly agreed upon road conduct~\cite{ViennaConv}, national rules~\cite{codes2015highway} and road traffic laws~\cite{RoadTraffic1988} which form a body of safe and legal driving rules, termed assertions, that must not be violated. 

Design confidence is gained when the autonomous vehicle (AV) can be shown to comply with these rules e.g.,\ through assertion checking during simulation. There have been several fatalities with AVs, some of which could be attributed to insufficient verification and validation (V\&V), e.g.~\cite{FatalityExample}. 

While on-road testing is an essential part of AV verification, it can be complemented by  simulation-based testing, which offers a means to explore the vast parameter space safely and efficiently~\cite{korosec2019waymo} whilst reducing the amount of costly road trials~\cite{kalra2016driving}.
%
%While on-road testing is an essential and complementary part of AV verification, simulation environments offer a means to explore the vast parameter space safely and efficiently~\cite{korosec2019waymo} whilst limiting the need for costly road trials~\cite{kalra2016driving}. 
In particular, simulations can be biased to increase the frequency at which otherwise rare events occur~\cite{Koopman2018}; this includes testing how the AV reacts to unexpected behaviour of the environment~\cite{RobustnessAutonomy}. 

Increasingly, the autonomous vehicle community is adopting game engines as simulation platforms to support the development and testing of vehicle control software. 
CARLA~\cite{CARLA_paper}% ~\cite{carla_main_website}
, for instance, is an open-source simulator for autonomous driving that is implemented in the Unreal Engine\footnote{\url{https://www.unrealengine.com/}}
% ~\cite{UE4_main_website}
, a real-time 3D creation environment for the gaming and film industry as well as other creative sectors. 

State-of-the-art game engines provide a convenient option for simulation-based testing. Strong arguments exist that they offer sufficient realism~\cite{Koopman2018} in the physical domain combined with realistic rendering of scenes, potentially suitable for perception stack testing and visual inspection of accidents or near misses. 
Furthermore, they are easy to set up and run compared to on-road testing and are simple to control and observe, both with respect to the environment the AV operates in as well as the temporal development of actors~\cite{Ulbrich2015}. 
Finally, support for hardware-in-the-loop development or a real-time test-bed for cyber-security testing~\cite{Javaid2013} may also be provided if required. 
Compared to the vehicle dynamics simulators and traffic-level simulators used by manufacturers~\cite{FrameworkAndChallenges}, game engines offer a simulation solution that meets many of the requirements for the development and functional safety testing of the autonomous features of AVs in simulation. 
However, while game engines are designed primarily for performance to achieve a good user experience, the requirements for AV verification go beyond that and include \textit{determinism}.

In this paper, we investigate non-determinism and how it affects simulation results using the example of CARLA, an open-source autonomous driving simulation environment based on the Unreal Engine.
In our case study, scenarios between pedestrian and vehicle actors are investigated to determine the actor position variance in the simulation output for repeated simulation runs. 
We establish that the CARLA simulator is non-deterministic.  Actor path variance was found to be non-zero and, under certain conditions, a deviation from the mean was observed of up to $59$cm.
In such an urban environment, we consider a deviation of up to $1$cm to be permissible for AV verification.
However, in our experiments, CARLA only exhibits this permissible variance when system utilisation is restricted to 75\% or less and the simulation is terminated once a vehicle collision has been detected.

The insights gained from this case study motivated the development of a general step-by-step method for AV developers and verification engineers to determine the simulation variance for a given simulation environment. 
Knowing the simulation variance will help assess the suitability of a game engine for AV simulation. In particular, this can give a better understanding of the effects of non-determinism and to what extent simulation precision may impact verification results.

This paper is structured as follows.
Section~\ref{s:prelim} defines terms used throughout the paper and identifies when determinism is needed. 
Section~\ref{s:background} briefly introduces how game engines work before investigating in Section~\ref{s:nondeterminisimSources} the potential sources of non-determinism in game engines.
An empirical case study of simulation variance for a number of scenarios involving pedestrians and vehicles is given in Section~\ref{s:case-study} including internal and external setting and system screening tests. 
The results from the case study are presented in Section~\ref{s:FinalResultsSection}. 
Section~\ref{s:methodology} presents the step-by-step method to assess the suitability of a simulation system for AV verification in general. 
We conclude in Section~\ref{s:conclusion} and give an outlook on future work.

\section{Preliminaries} \label{s:prelim}
\subsection{Definitions}
Refer to Fig.~\ref{variance_description} for the definitions introduced in this section.
% Several definitions are introduced in this section. These are used in the subsequent discussion. Refer to Fig.~\ref{variance_description} throughout this section.
\\

\subsubsection{Determinism}

Schumann et al.\ describe determinism as the property of causality given a temporal development of events such that any state is completely determined by prior states~\cite{Schumann2010}. However, in the context of simulation this should be expanded to include not just prior states but also the history of actions taken by all actors. Therefore, a deterministic simulation will always produce the same result given the same history of prior states and actions.

\begin{figure}[!t]
    \centering

    \begin{subfigure}{.48\textwidth}
        \includegraphics[width=1\textwidth]{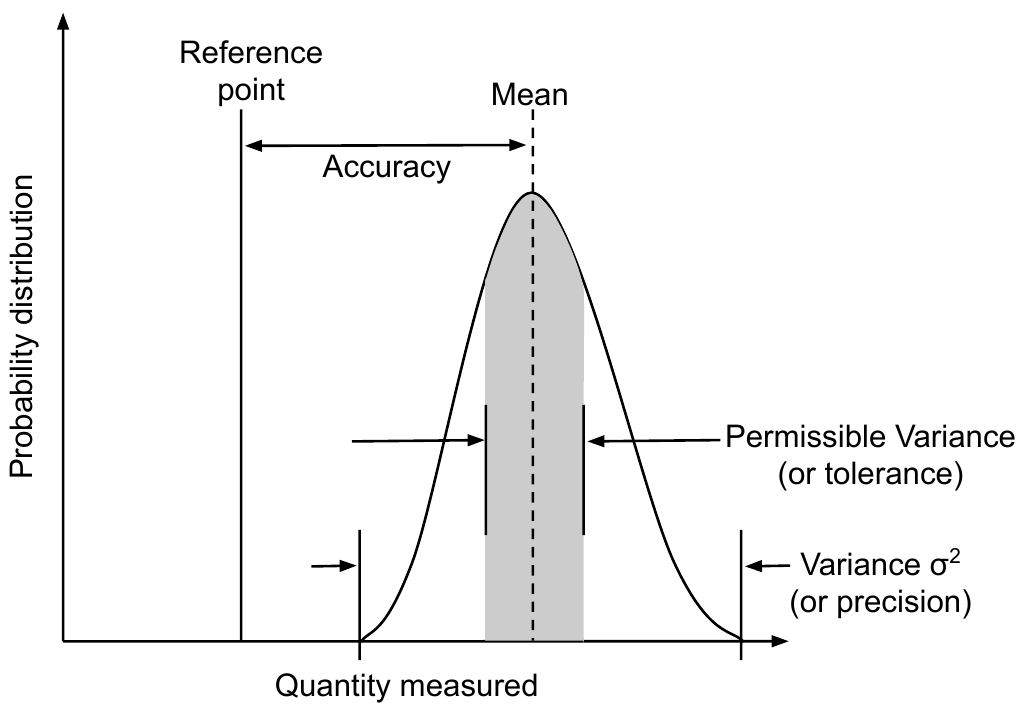}
        \caption{Non-determinism}
        \label{variance_description_a}
    \end{subfigure}

    \begin{subfigure}{.48\textwidth}
        \includegraphics[width=1\textwidth]{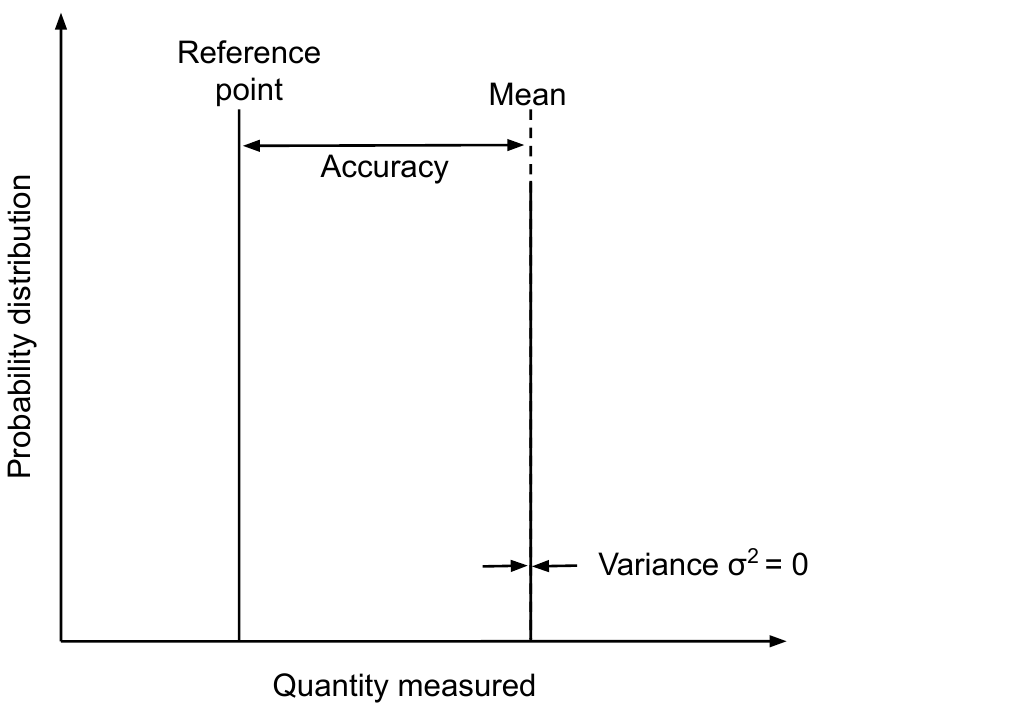}
        \caption{Determinism}
        \label{variance_description_b}
    \end{subfigure}

    \caption{Illustration of accuracy, variance, precision and tolerance in the context of non-determinism and determinism}
    \label{variance_description}
\end{figure}

A simulation can be thought of as the process of generating or producing experimental data. In the case of a driving simulator, kinematics will describe future states of actors given the current conditions and actions taken, thereby generating new data. If a simulation is deterministic, Fig.~\ref{variance_description_b}, then there will be no variation in the generated output data, i.e. all future states are perfectly reproducible from prior states and actions. However, if a simulation is non-deterministic, Fig.~\ref{variance_description_a}, then there will be a variation in the output data. \\

\subsubsection{Variance, Precision \& Tolerance}
We adopt terminology from the mechanical engineering and statistics domains to describe when there is variation in the generated output data~\cite{ADictionaryofMechanicalEngineering}. \textit{Variance} is used here to define the spread or distribution, of the generated output data with respect to the mean value. \textit{Precision} is synonymous with \textit{variance} although inversely related mathematically. Therefore, variance can indicate the degree to which a simulation can repeatedly generate the same result when executed under the same conditions and actions. \textit{Tolerance} is defined as the permissible limit of the variance, or in short the \textit{permissible variance}. 

As an analogy, the simulator can be thought of as a manufacturing process that produces data. To determine the precision of this process, the output must be measured and analysed for differences when the process is repeated. Those differences describe the spread or variance in the process output. A hard limit on the variance can then be defined, Fig.~\ref{variance_description_a}, beyond which the output fails to meet the required tolerance, e.g.\ the output is rejected by quality control. Real manufacturing fails to achieve absolute precision. Hence, there is a need for tolerances to be specified to account for the variance in real-world manufacturing processes. 

If a simulator is deterministic then it will produce results with absolute precision or zero variance, Fig.~\ref{variance_description_b}, and hence will be within tolerance by design. If the simulator is non-deterministic then there will be a measurable, non-zero variance in the output data.\\

\subsubsection{Accuracy}
Precision and tolerance should not be confused with \textit{accuracy}, which describes how closely the mean of the generated output of a process aligns to a known standard or reference value. Therefore, we define accuracy as the difference between the true value, or reference value, and what has been achieved in the data generation process or simulation. 
For a driving simulation, the reference value may be the real world that the simulation seeks to emulate, where any divergence from this standard is termed the \textit{reality gap}. 
In practice, full accuracy will often not be achievable due to modelling and associated computational demands of creating and executing an exact replica. In most cases it is unnecessary and some authors state that `just the right amount of realism' is required to achieve valid simulation results~\cite{Koopman2018}. \\

\subsubsection{Simulation Trace}
A simulation trace is the output log from the simulator consisting of a time series of all actor positions ($x,y,z$) in a 3D environment recorded at regular time intervals. This definition could be extended to include other variables. A set of simulation traces derived from the same input and starting state then forms the experimental data on which variance is calculated for a given simulation run. \\

\subsubsection{Simulation Variance \& Deviation}
If the simulator is non-deterministic then how can the simulation variance be measured? This can be achieved by monitoring the values of any of the recorded output variables that should be consistent from run to run. For example, actor position variance is a distance-based metric that can be derived from simulation traces. The actor position over time, i.e.\ the actor path, is often used in assertion checking, e.g.\ to determine whether vehicles keep within lanes or whether minimum distances to other vehicles and road users are being maintained. 
Thus, in the case study presented in this paper, the term \textit{simulation variance}, measured in SI unit $m^2$, refers to a measure of actor path variance in the simulation, assuming fixed actions. Case study results are presented using deviation (SI unit $m$), the square root of variance, rather than variance, as this is a more intuitive measure to comprehend when interpreting test results.
\\

\subsubsection{Scene, Scenario \& Situation}
We adopt the terminology defined for automated driving in~\cite{Ulbrich2015}, where \textit{scene} refers to all static objects including the road network, street furniture, environment conditions and a snapshot of any dynamic elements. Dynamic elements are the elements in a scene whose actions or behaviour may change over time; these are considered actors and may include the AV, or \textit{ego vehicle}, other road vehicles, cyclists, pedestrians and traffic signals. The \textit{scenario} is then defined as a temporal development between several scenes which may be specified by specific parameters. A \textit{situation} is defined as the subjective conditions and determinants for behaviour at a particular point in time.

\subsection{When is Determinism needed?}

% NEW TEXT
Determinism is a key requirement for simulation during AV development and testing, it ensures repeated runs have the same output and therefore have zero variance. A simulator with non-zero variance is non-deterministic but may be sufficient for some applications as long as variance is permissible, i.e. \textit{within tolerance}. Therefore, \textit{tolerance} is the acceptable degree of variability between repeated simulations. When the simulation output is within tolerance, coverage results are stable and, when a test fails, debugging can rely on the test producing the same trace and outcome when repeated. This ensures that software bugs can be found and fixed efficiently, and that simulation results are trustworthy.

Non-deterministic simulation may have non-zero variance in, for example, actor positions, which may render the outcome unstable producing incorrect results potentially leading false confidence in the system under test. 
%in intermittent assertion failure, failing to detect bugs and potentially leading to false confidence in the system under test. 
When used for gaming, game engines do not need to be deterministic nor do they have any requirements on the limits of permissible variance; there are no safety implications from non-determinism in this domain, nor is finding and fixing all the bugs a high priority for games developers. It could even be argued that simulation variance is a feature that enhances gaming and improves the user experience. However, the situation is very different for AV development and testing. Thus, our main research questions are:
{\em How can one assess whether a simulation environment is deterministic?} and 
{\em How can one determine and control the simulation variance?}

\section{Background} \label{s:background}

There are numerous game engines with their associated development environments that could be considered suitable for AV development, e.g.\ Unreal Engine
, Unity\footnote{\url{https://unity.com/}} and CryEngine.\footnote{\url{https://www.cryengine.com/}} 
% ~\cite{CryEngine_main_website}
Specific autonomous driving research tools have been created to abstract and simplify the development environment, some of which are based on existing game engines, e.g.\ CARLA~\cite{CARLA_paper}, AirSim\footnote{\url{https://microsoft.github.io/AirSim/}}, Apollo\footnote{\url{http://apollo.auto/}}, and some have been developed for cloud-based simulation, e.g.\ Nvidia Drive Constellation\footnote{\url{https://www.nvidia.com/en-gb/self-driving-cars/drive-constellation/}}.
%~\cite{UE4_main_website}
% ~\cite{Unity_main_website}
%~\cite{carla_main_website}
% ~\cite{AirSim_main_website}
% ~\cite{Apollo_main_website}
% ~\cite{nvidia_constellation}

Investigating the determinism of game engines has not attracted much research interest to date since performance is more critical for game developers than accurate and repeatable execution. Ensuring software operates deterministically is a non-trivial task.  Catching intermittent failures, or flaky tests~\cite{intermittently-failing-tests}, in a test suite that cannot be replayed makes the debugging process equally difficult~\cite{acm-q-rr-interview}. This section gives an overview of the internal structure of a game engine and what sources or settings may affect \textit{simulation variance}.

\begin{figure}[h]
\centering
\includegraphics[width=0.5\textwidth]{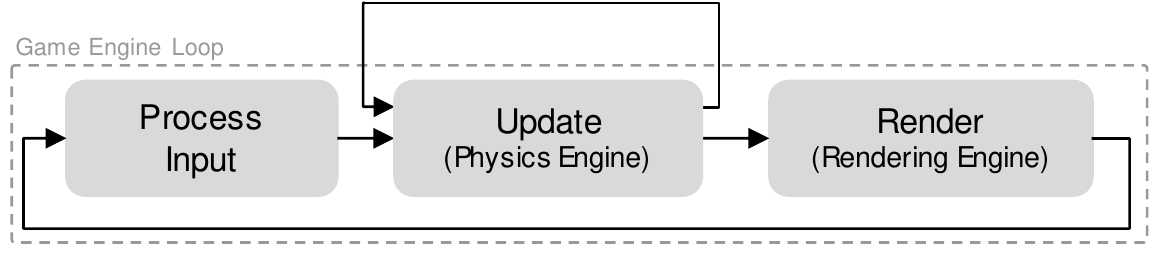}
\caption{Game engine loop block diagram~\cite{GameProgPatternsBook}.}
\label{GameEngineLoopDiagram}
\end{figure}

% ======= Overview of a Game Loop
Central to a game engine are the main game logic, the artificial intelligence (AI) component, the audio engine, and the physics and rendering engines. For AV simulation, we focus on the latter two. The game loop is responsible for the interaction between the physics and rendering engines. Fig.~\ref{GameEngineLoopDiagram} depicts a simplified representation of the process flow in a game engine loop, where  initialisation, game logic and decommissioning have been removed\footnote{\url{https://docs.unity3d.com/Manual/ExecutionOrder.html}}. 
% ~\cite{unity_event_execution}.
A game loop is broken up into three distinct phases: processing the inputs, updating the game world (Physics Engine), and generating outputs (Rendering)~\cite{GameEngineArchBook}.

The game loop cycle starts with initialising the scene and actors. Input events from the User or AI are then processed followed by a physics cycle which may repeat more than once per rendered frame if the physics time step, $dt$, is less than the render update rate. This is illustrated by the loop in the physics update in Fig.~\ref{GameEngineLoopDiagram}. The render update will process frames as fast as the computational processing will allow up to the maximum monitor refresh rate~\cite{unity_framerates}. When the frame is rendered the game loop cycle returns to processing inputs. For an intuitive and  more detailed description of the interplay between the physics and render cycles see~\cite{JohnAustinUnity}. 
% \footnote{\url{https://johnaustin.io/articles/2019/fix-your-unity-timestep}}.% is given 

The physics engine operates according to a time step, $dt$. The shorter this time step is, the smoother the interpretation of the physical dynamics will be. To use a fixed physics time step, the user's display refresh rate needs to be known in advance. This requires an update loop to take less than one render tick (one frame of real world time). Given the range of different hardware capabilities, a variable delta time is often implemented for game playing, taking the previous frame time as the next $dt$. However, variable $dt$ can lead to different outcomes in repeated tests and in some cases unrealistic physical representations\footnote{\url{https://gafferongames.com/post/fix_your_timestep/}}. %~\cite{gaffer}. 
Semi-fixed or limited frame rates ensure $dt$ does not exceed some user-defined limit to meet a minimum standard of physical representation but allow computational headroom for slower hardware. Some engines provide sub-stepping which processes multiple physics calculations per frame at a greater CPU cost, e.g.\ Unreal Engine~\cite{UE4_substepping}. If the engine tries to render between physics updates, \textit{residual lag} can occur, which may result in frames arriving with a delay to the simulated physics. Thus, extrapolation between frames may need to be performed to smooth transition between scenes. Note that both residual lag and extrapolation could affect perception stack testing. In exceptional cases, where computational resources are scarce, the fixed time step can be greater than the time between render ticks and the simulation will exhibit lag between input commands and rendered states, resulting in unsynchronised and unrealistic behaviour as can be experienced when games are executed on platforms not intended for gaming. 

Considering the objectives for gaming and comparing them to those for AV development and testing, there are fundamental differences. Providing game players with a responsive real-time experience is often achieved at the cost of simulation accuracy and precision.
The gamer neither needs a faithful representation of reality (i.e. gamer accepts low accuracy) nor requires repeated actions to result in the same outcome (i.e. gamer accepts low precision). In contrast, high accuracy and precision are necessary for AV development, testing and verification.

% ======= Sources of non-determinism
\section{Potential Sources of Non-Determinism} \label{s:nondeterminisimSources}

The following review discusses the potential sources of non-determinism that were found in the literature or found as part of our investigation into game engines. We have examined hardware- as well as software-borne sources of non-determinism that occur at different layers of abstraction. 
A good analysis of potential sources is given by Strandberg et al.~\cite{intermittently-failing-tests}, although the AV simulation domain introduces its own unique challenges that were not considered in that paper.

\medskip

% ======= Floating-Point Numbers
\subsection{Floating-Point Arithmetic}

% NEW TEXT V3
It is a common misconception to attribute non-deterministic computational execution with the use of floating-point number representation, which necessitates rounding due to a fixed bit width~\cite{FloatingPointsBook,goldberg1991every}. 
% available and can potentially result in rounding errors~\cite{FloatingPointsBook,goldberg1991every}. 
As a consequence, floating-point arithmetic is not associative~\cite{Kapre2007} and results may differ depending on execution order.
% Some authors even suggest avoiding floating point representation entirely~\cite{empirical-analysis-of-flaky-tests}. 
%
In the context of AV simulation, this
% finite-precision of floating point number representation 
% such rounding errors 
could result in accuracy issues of, for example, actor positions. While some authors suggest
avoiding floating-point representation entirely~\cite{empirical-analysis-of-flaky-tests}, we argue that the precision issues related to floating-point operations are better described as \textit{incorrectness} that is in fact repeatable; they do not cause non-determinism per se. So, even if the result of a mathematical operation is incorrect due to floating-point rounding errors, it should always be equally \textit{incorrect} when repeated for implementations that meet the IEEE floating-point standard~\cite{8766229}. 
However, different compiler configurations, aggressive optimisations~\cite{llvm-floating-point}, 
parallelisation within the runtime environment or at the hardware level and performing the execution on a GPU rather than a CPU\footnote{ CPU and GPU processors may have different register widths~\cite{Whitehead2011} } may all affect the execution order. 
In conclusion, floating-point arithmetic does not cause non-zero \textit{simulation variance} for repeated simulation runs when using the same executable, hardware, configuration and execution order.

\subsection{Scheduling, Concurrency and Parallelisation}

Runtime scheduling is a resource management method for sharing computational resources between tasks of different or equal priority depending on the operating system's scheduler policy. A scheduler policy may be optimised in many ways such as for task throughput, deadline delivery or minimum latency~\cite{liu1973scheduling}. In principle, changing the scheduling policy and thread priorities may increase simulation variance. 
It would therefore be important to ensure these remain stable between repeated runs. 
% Changes to thread priorities or scheduling policy during repeated runs on the same hardware and should therefore give the same output. 
However, if some aspects of the game loop are multi-threaded\footnote{\url{https://docs.unity3d.com/Manual/JobSystemMultithreading.html}} 
% ~\cite{unity_multithreading} 
or if the scheduler simply randomly selects from a set of threads with equal priority, this may alter an otherwise deterministic sequence of events. 

Similar to thread scheduling, scheduling at the hardware level on a multi-core system determines on which processor core to execute processes. This may be decided based on factors such as throughput, latency or CPU utilisation. Scheduling multiple processes across several processing cores, where the number of cores is smaller than the number of processes can result in variation of the execution order and cause simulation variance unless explicitly constrained or statically allocated prior to execution. Indeed, the developers of the debugging program \texttt{rr}~\cite{RR_link} took significant steps to ensure deterministic behaviour of their program by executing or context-switching all processes to a single core, which avoids data races as single threads cannot concurrently access shared memory. This allowed control over the scheduling and execution order of threads, thus promoting deterministic behaviour by design~\cite{acm-q-rr-interview}. 
Likewise, simulation variance may be observed for game engines that use GPU parallelisation  by offloading time-critical calculations to several dedicated cores simultaneously. While this would be faster than a serial execution, the order of execution arising from program-level concurrency cannot, in general, be guaranteed. 

Overall, scheduling, concurrency and parallelisation may be reasons for \textit{simulation variance}.

\subsection{Non-Uniform Memory Access (NUMA)} \label{s:sources_numa}
For a repeated test that operates over a number of cores based on a CPU scheduling policy, memory access time may vary depending on the physical memory location relative to the processor. Typically a core can access its own memory with lower latency than that of another core resulting in lower interprocessor data transfer cost~\cite{nieplocha1996global}. 
Changes in latency between repeated tests may, in the worst case, cause the game engine to operate non-deterministically if tasks are processed out of sequence using equal priority scheduling, or, perhaps, simply with an increased data transfer cost, i.e.\ slower. 
By binding a process to a specific core for the duration of its execution, the variations in data transfer time can be minimised.

\subsection{Error Correcting Code (ECC) Memory}
ECC Memory is used ubiquitously in commercial simulation facilities and servers to detect and correct single bit errors in DRAM memory~\cite{Dell1997}. Single bit errors may occur due to malfunctioning hardware, ionising radiation (background cosmic or environmental sources) or from electromagnetic radiation~\cite{dodd2003basic}. If single bit errors go uncorrected then subsequent computational processing will produce incorrect results, potentially giving rise to non-determinism due to the probabilistic nature of such errors occurring. Estimating the rate of error is difficult and dependent on hardware, environment and computer cycles~\cite{mielke2008bit}.

Any simulation hardware not using ECC memory that runs for 1000's of hours, typical in AV verification, is likely to incur significant CPU hours and is therefore subject to increased exposure to these errors. To counter this, commercial HPC and simulation facilities typically employ ECC memory as standard.

\subsection{Game Engine Setup}
The type and version of the engine code executed should be considered, paying attention to the control of pseudo-random numbers, fixed physics calculation steps, ($dt$), fixed actor navigation mesh, deterministic ego vehicle controllers and engine texture loading rates especially for perception stack testing. 
For example, in Unreal Editor the \textit{unit}~\cite{stat_commands} command can be used to monitor performance metrics such as \textit{Frame} which reports the total time spent generating one frame, \textit{Game} for game loop execution time and \textit{Draw} for render thread time. 
With respect to perception stack testing, weather and lighting conditions in the game engine should be controlled as well as any other dynamic elements to the simulation environment, e.g.\ reflections from surface water, ensuring textures are not randomly generated.

\subsection{Actor Navigation}

The A* algorithm is commonly used for actor navigation in game engines.
A* explores a search graph through a series of node-selection-and-expansion steps that continue until a goal node is found.
A heuristic is used at each step to select the most promising node for expansion from a set of candidate nodes called the \textit{frontier}.
The heuristic is not required to guarantee a uniquely preferred node in the frontier, so ties may be found during selection and then broken arbitrarily, meaning that an implementation of A* can be non-deterministic.

For reasons of efficiency, A* represents the frontier as a priority queue with nodes prioritised by the heuristic.
Selection then reduces to a dequeue operation on the frontier.
How to break ties in priority queues is related to a broader question of sorting stability~\cite{Sedgewick2011} where a priority queue is stable if it breaks ties based on insertion order, and is unstable otherwise.
Typically unstable priority queues are more computationally efficient than stable priority queues and the most common method (using a binary heap)~\cite{boost2021,priorityQueue} is unstable, breaking ties based on internal heap order rather than insertion order.
While this priority queue is unstable, it is still deterministic, since it always breaks ties in the same way (based on heap order).
An example of an unstable priority queue that is non-deterministic is one that breaks ties at random.
An implementation of A* is deterministic if it uses either a stable priority queue or an unstable but deterministic priority queue.

The Unreal documentation and source code suggest that CARLA uses a priority queue based on a binary heap 
% e.g. see\footnote{\url{https://docs.unrealengine.com/en-US/API/Runtime/AIModule/FGraphAStar/index.html}, \url{https://github.com/EpicGames/UnrealEngine/blob/release/Engine/Source/Runtime/AIModule/Public/GraphAStar.h} and \url{https://docs.unrealengine.com/en-US/ProgrammingAndScripting/ProgrammingWithCPP/UnrealArchitecture/TArrays/index.html}}.
~\cite{FGraphAStar,TArray,GraphAStar}. 
Theoretically this means that use of A* in CARLA will be deterministic as long as the heap is always constructed in the same way.
However, if some unknown change causes the heap to be constructed in a different way, then the use of A* may appear non-deterministic.
This can be a complex problem in practice.
For example, suppose A* iterates over actions in order of hash value, but between runs there is some untracked change to the hardware/software stack that alters the hash function, then this may change insertion order of nodes to the frontier, which could eventually lead to a different binary heap and thus a different optimal solution found by A*.
This suggests that while a simulated environment may behave deterministically, factors outside the simulator may cause changes to the operation of an A* implementation, which would then materialise as non-deterministic runs in the simulator.
The difference between use of a stable priority queue and an unstable but deterministic priority is simply that it may be easier to detect such changes under the former because insertion order is typically more meaningful with respect to the actual implementation of A*.

\subsection{Summary}
We have investigated the potential sources of non-determinism affecting game engines and explored the impact they may have on simulation variance. Memory checking not withstanding, errors associated with the lack of ECC are likely to be minimal unless there is significant background radiation or 1000's of hours of computation are expected. To ensure precise simulation outcomes, the physics setting, $dt$, must be fixed, along with any actor navigation meshes, seeds for random number generation, game engine setup and simulation specific parameters. %
Any implementation of the A* search algorithm for actor navigation must use a stable priority queue to ensure deterministic results. 
Non-uniform memory access (NUMA) should only affect interprocessor data transfer cost and, without control measures, will only make the computation cycle longer. Relative access times between different caches are likely to be small although may have a more pronounced impact on high throughput systems, e.g.\ HPC. If this change in computational cycle gives opportunity for the execution order to be changed then this situation may lead to non-determinism. 

Basic thread scheduling should not affect the simulation's determinism unless changing scheduling policy, operating system or migrating between machines with different setups. However, should new and unexpected threads start during the simulation, then the interruption to execution order or additional resource demand may affect timing of subsequent steps, thus reducing the number of physics updates within a game loop. Likewise, uncontrolled allocation of hardware resources such as CPUs or GPUs can potentially give rise to non-determinism.

% ***************************************************
%  Empirical Investigation
% ***************************************************
\section{Case Study of Simulation Variance} \label{s:case-study}

We present an empirical investigation into using game engines for
simulation-based verification of autonomous vehicles with a focus on
characterising sources of non-determinism in order to understand  the impact
they have on simulation variance. 
Gao et al.~\cite{when-and-what-should-we-control} took a similar approach investigating Java applications, where a set of sources of non-determinism (termed factors) were shown to impact on repeatability of testing. Ultimately, our objective is to control non-determinism to minimise simulation variance.

We first describe the context, scene and scenario of interest before discussing and defining a tolerance for what is considered an acceptable simulation variance in this context. A discussion on the internal and external settings of the simulation is included, along with system configuration and pre-screening sections. 

% ***************************************************
%  Scenario Description
% ***************************************************
\subsection{Context, Scene and Scenario}\label{TestsDescriptionAndTechnicalities}
\begin{figure}[!t]
    \centering
    \begin{subfigure}{.24\textwidth}
        \includegraphics[width=1\textwidth]{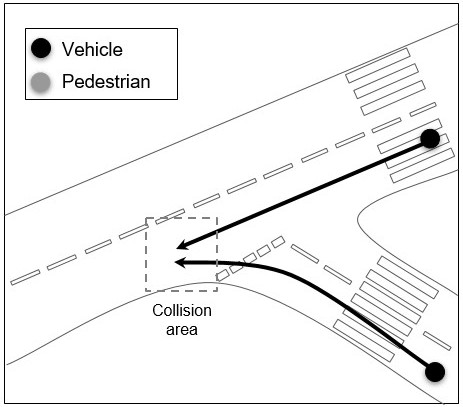}
        \caption{}
        \label{Test_a}
    \end{subfigure}
    \begin{subfigure}{.24\textwidth}
        \includegraphics[width=1\textwidth]{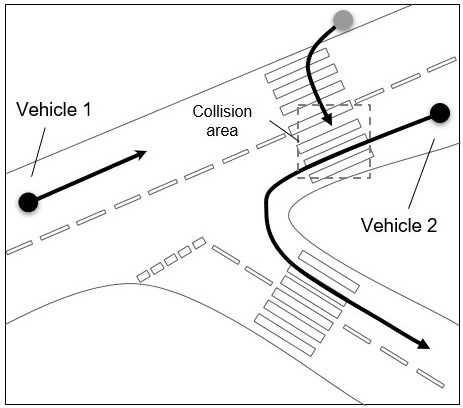}
        \caption{}
        \label{Test_b}
    \end{subfigure}
    \begin{subfigure}{.24\textwidth}
        \includegraphics[width=1\textwidth]{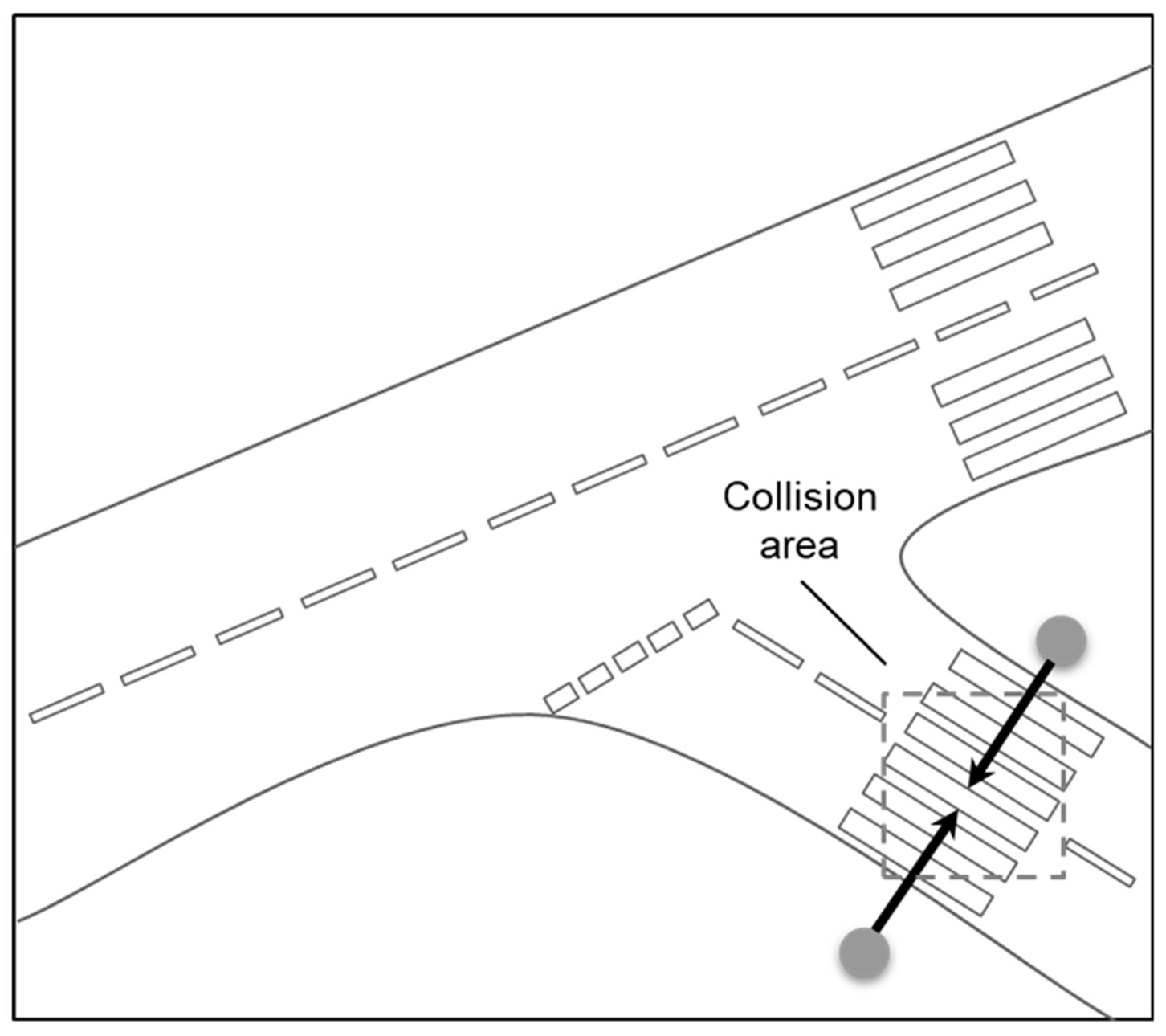}
        \caption{}
        \label{Test_c}
    \end{subfigure}
    \caption{Schematic of test scenarios for (a) Tests 1-2, (b) Tests 3-4, (c) Tests 5-6. Descriptions are given in Table~\ref{TableOfExperiments}.}\label{f:test_a_and_b}
\end{figure}

This case study draws on a setup used to verify an urban mobility and transport solution, where the primary verification objective is to test the behaviour of an ego vehicle in an urban environment at a T-junction in response to pedestrians and other vehicles. Thus, the scene for our investigation is the T-junction with the scenarios as shown in Figure~\ref{f:test_a_and_b}.

This scene was used to create a number of scenarios involving pedestrians and vehicles in order to identify any changes in the actor paths over repeated tests executed under a variety of systematically designed conditions and hence study any simulation variance. The vehicles and pedestrians were given trajectories, via pre-defined waypoints, that would result in either colliding with or avoiding other actors.

\subsection{Tolerable Simulation Variance}\label{s:tolerance}
To achieve stable verification results over repeated test runs, the simulated actor states must be precise to a specific tolerance. Deterministic behaviour would result in zero variance of the simulated actor states but if this cannot be achieved then what is permissible? The tolerance must be appropriate to allow accurate assertion checking and coverage collection in the simulation environment, but not so small such that assertion checking would fail with minor computational perturbations. Thus, a tolerance must be defined to reflect the precision at which repeatability of simulation execution is required. 

For this case study a tolerance on actor position of $1$m would be insufficient when considering the spacial resolution required to distinguish between a collision and a near-miss event. A very small value, e.g.\ $1\times10^{-12}m$, may be overly-sensitive to minor computational perturbations and generate false positives. Therefore, for this case study a tolerance of $1$cm has been selected and thus any variance of less than $1$cm is permissible. To put this another way, we can accept a precision with a tolerance of $\leq$$\pm$1$cm$. In practice, this tolerance may need to be chosen for each specific verification case and the speed of the vehicles within the environment~\cite{jiang2014intercultural}.

In the following, case study results are shown in terms of the maximum deviation, $\max\sigma$, from the mean actor path over the entire simulation history where any value higher than the specified tolerance is considered non-permissible.

% ***************************************************
%  Actor Collisions
% ***************************************************

\begin{table*}[t]
\centering
\begin{tabular}{clclccc}
\toprule
Test & Actors               & Collision    & Collision Type        & $n$  & $\max\sigma$ (m) & $\max\sigma$ (m) \\ 
   &                  &        &               &      & (unrestricted)  & (restricted)  \\ \midrule
1    & Two vehicles                   & No       & N/A               & 1000 & 0.03 & $7.0{\times}10^{-3}$ \\
2    & Two vehicles                   & Yes      & Vehicle and Vehicle     & 1000 & 0.31 & $9.8{\times}10^{-3}$ \\
3    & Two vehicles and a pedestrian  & No       & N/A               & 1000 & 0.07 & $5.2{\times}10^{-4}$ \\
4    & Two vehicles and a pedestrian  & Yes      & Vehicle and Pedestrian    & 1000 & 0.59 & $1.5{\times}10^{-12}$ \\
5    & Two pedestrians                & No       & N/A               & 1000 & $5.6{\times}10^{-13}$ & $5.6{\times}10^{-13}$ \\
6    & Two pedestrians                & Yes      & Pedestrian and Pedestrian & 1000 & $5.6{\times}10^{-13}$ & $5.6{\times}10^{-13}$ \\
\bottomrule
\end{tabular}
\caption{A description of the test scenarios showing the test number, the actors included, whether a collision occurred and if so then between which actors. $n$, the number of repeats is set to 1000 and $\max\sigma$ is the maximum simulation deviation. The term \textit{unrestricted} refers to an unrestricted account of the results including results of any resource utilisation. To understand the impact of collisions and high resource utilisation, the \textit{restricted} column shows a subset of the results where post-collision data and experiments above 75\% resource utilisation have been removed.}
\label{TableOfExperiments}
\end{table*}

\subsection{Actor Collisions}\label{S:Actor_Collisions}
Previous investigations into the Unreal Engine indicated that collisions between actors and solid objects, termed \textit{blocking physics bodies} in Unreal Engine documentation~\cite{collision_overview}, can lead to high simulation variance~\cite{TSLUnrealEngineTesting}. 
Collisions and the subsequent physics calculations that are processed, termed \textit{event hit} callback in Unreal Engine, were identified as potentially key aspects to the investigation into simulation variance.

The tests used for the case study are listed in Table~\ref{TableOfExperiments}. They cover a range of interactions between actor types. The map size for the simulation test environment was 354$m$ $\times$ 170$m$. The range of movement of the actors within the environment was up to 70$m$ for vehicles and up to 25$m$ for pedestrians.
Tests 1 \& 2 involve two vehicles meeting at a junction where they either do not collide (Test 1) and where they do collide (Test 2), thereby triggering an \textit{event hit} callback in the game engine. In both cases the trajectories of the vehicles are hard-coded to follow a set of waypoints spaced at $0.1$m intervals using a PID controller. 
In Test 3 a mixture of different actor types is introduced where two vehicles drive without collision and a pedestrian walks across the road at a crossing point. 

Similar to vehicles, pedestrian actors navigate via a set of regularly spaced waypoints at $0.1$m intervals using the A* search algorithm which is the default method to find optimal paths for the CARLA pedestrian actors~\cite{newton2016unreal}. 
There is evidence to suggest that this actor navigation in CARLA could be a source of non-deterministic simulation behaviour~\cite{CARLABenchmark}. 
This behaviour is explored in Test 4 where a pedestrian collides with one of the vehicles at the crossing, triggering an \textit{event hit} callback, see Fig~\ref{Test_a}.  
%
% Unreal Engine is the underlying framework for the CARLA simulator which uses the A* search algorithm to find optimal paths for actor navigation~\cite{newton2016unreal}.

Tests 5 \& 6 involve only pedestrians that either, do not collide (Test 5) and that do collide (Test 6), see Fig.~\ref{Test_b}.

% ***************************************************
%  Experiment Description
% ***************************************************
\subsection{Evaluation Metric}\label{s:Experiment_Description}
For each test the position of each actor is logged at $0.1s$ intervals providing a trace of that actor's path with respect to simulation time. The logs from repeated tests are sourced to establish a value for the variance associated with each actor, $a$, at each time point $t$, giving a variance function over time for each actor, $\sigma_a^{2}(t)$.

Instead of using variance, herein the results are given in terms of the deviation, $\sigma_a(t)$, which indicates the dispersion of the actor path relative to the mean and is helpfully in the same units as actor position, i.e.\ metres ($m$), for ease of interpretation. The maximum variance ($\Psi$) over the entire set of $n$ repeated simulations, i.e.\ the overall observed worst case, is defined as the largest variance of any actor at any time in any of the simulation runs, as given in Equation~\ref{eq:max_sigma}. 
\begin{equation} \label{eq:max_sigma}
\Psi = \max_{a,t}\sigma_a^{2}(t)
\end{equation}
The maximum deviation is the absolute value of the square root of the maximum variance and herein referred to as ${\max\sigma}$ for brevity. 

The maximum deviation, $\max\sigma$, can be analysed for the different scenarios and settings that were identified as potential sources of non-determinism, and compared against the limit of \textit{permissible variance} to indicate if the simulation is sufficiently accurate for verification purposes.

% ***************************************************
%  Internal Settings
% ***************************************************
\subsection{Simulator Settings}

Within Unreal Engine there are numerous internal settings relating to the movement and interaction of physical bodies in the simulation. Settings can be adjusted to alter how actors interact and path plan via the navigation mesh of the environment, e.g.\ \textit{Contact Offset} and \textit{Navmesh Voxel Size}, or can be changed to improve the fidelity of physics calculations between game update steps, e.g.\ \textit{Physics Sub-Stepping} and \textit{Max Physics Delta Time}. Other options such as \textit{Enable Enhanced Determinism} were investigated along with running the engine from the command line with options for more deterministic behaviour \texttt{-deterministic}, floating-point control \texttt{/fp:strict} and headless mode \texttt{-nullrhi} along with running the test as a packaged release by building and cooking~\cite{releasing_project}. An initial study into the Unreal Engine using a pedestrian and a moving block was used to investigate simulation variance against these settings. The results were compared to a baseline of the default engine settings. However, none of these options improved simulation variance significantly and all internal setting were set restored to the default values. Details on this previous investigation can be found on the Trustworthy Systems github~\cite{TSLUnrealEngineTesting}.

% ***************************************************
%  External Settings
% ***************************************************
\subsection{External Settings}

Executing physics calculations during simulation may consume a significant proportion of system resources (e.g.\ CPU and GPU processors). This suggests that as resource utilisation of the simulation increases, so does the \textit{simulation variance} which was corroborated with some initial investigations~\cite{TSLUnrealEngineTesting} and explored more fully in this work using CARLA. 

% PREVIOUS TEXT V2
% After examining controls the internal game engine, external settings were explored. Game engines, and simulation hardware more generally, will utilise available system resources such as central and graphical processing units (CPU and GPU respectively), to perform physics calculations and run the simulation environment. For high performance simulations the demand on these resources may be a significant fraction of the total available and an initial hypothesis was that as this ratio tended toward one, there would be an increase in \textit{simulation variance}.

% Some initial exploratory work was undertaken~\cite{TSLUnrealEngineTesting} that suggested computational resource utilisation was positively correlated with simulation variance, leading to high simulation variance when a system is under high load. This initial observation was explored more fully in this work using the CARLA platform. 

To replicate in a controlled manner the high computational loads that may be anticipated for high performance simulations, software that artificially utilises resources on both the CPU and GPU were executed alongside the simulation. Resource utilisation was artificially increased for both CPU and GPU devices to include a range of values from 0 to 95\% (see Section~\ref{s:methodology}) using reported values of the system monitors \texttt{htop} and \texttt{nvidia-smi} respectively. Resource utilisation figures reported here should be considered approximate values. Resource utilisation was capped to 75\% in some parts of the results and referred to as \textit{restricted}. This was done to limit the load on the test system and hence limiting simulation variance. The term \textit{unrestricted} places no such limit on the system load.

Practitioners should also be aware that many libraries for calculating variance itself may require attention to get precise results. For example the \texttt{numpy} method of variance is sensitive to the input precision and will return an incorrect answer if the wrong parameters are set~\cite{NumpyVar}. In \texttt{matlab}, the calculation of variance may switch from single thread to a multi-threaded execution not obviously apparent to the user when the input data size becomes large enough, opening up the potential for concurrency-induced imprecision~\cite{matlab_parallel_computing}.

% ***************************************************
%  Results and Discussion
% ***************************************************

\section{Results and Discussion}\label{s:FinalResultsSection}
\begin{figure}[t]
    \centering
    \includegraphics[width=0.48\textwidth]{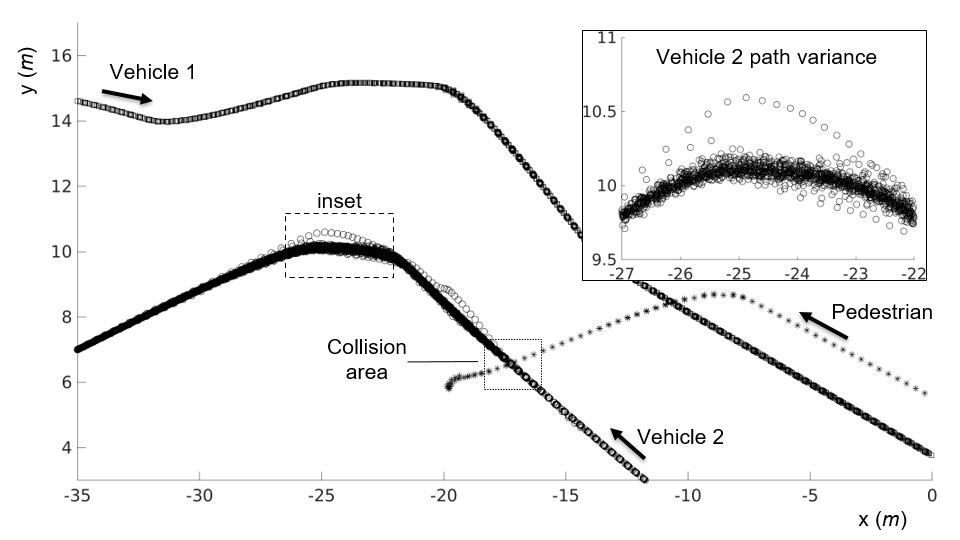}
    \caption{Actor path plot in X--Y plane for Test 4 with 95\% resource utilisation.}
    \label{actorPathPlot}
\end{figure}

\begin{figure}[t]
    \centering
    \includegraphics[width=0.48\textwidth]{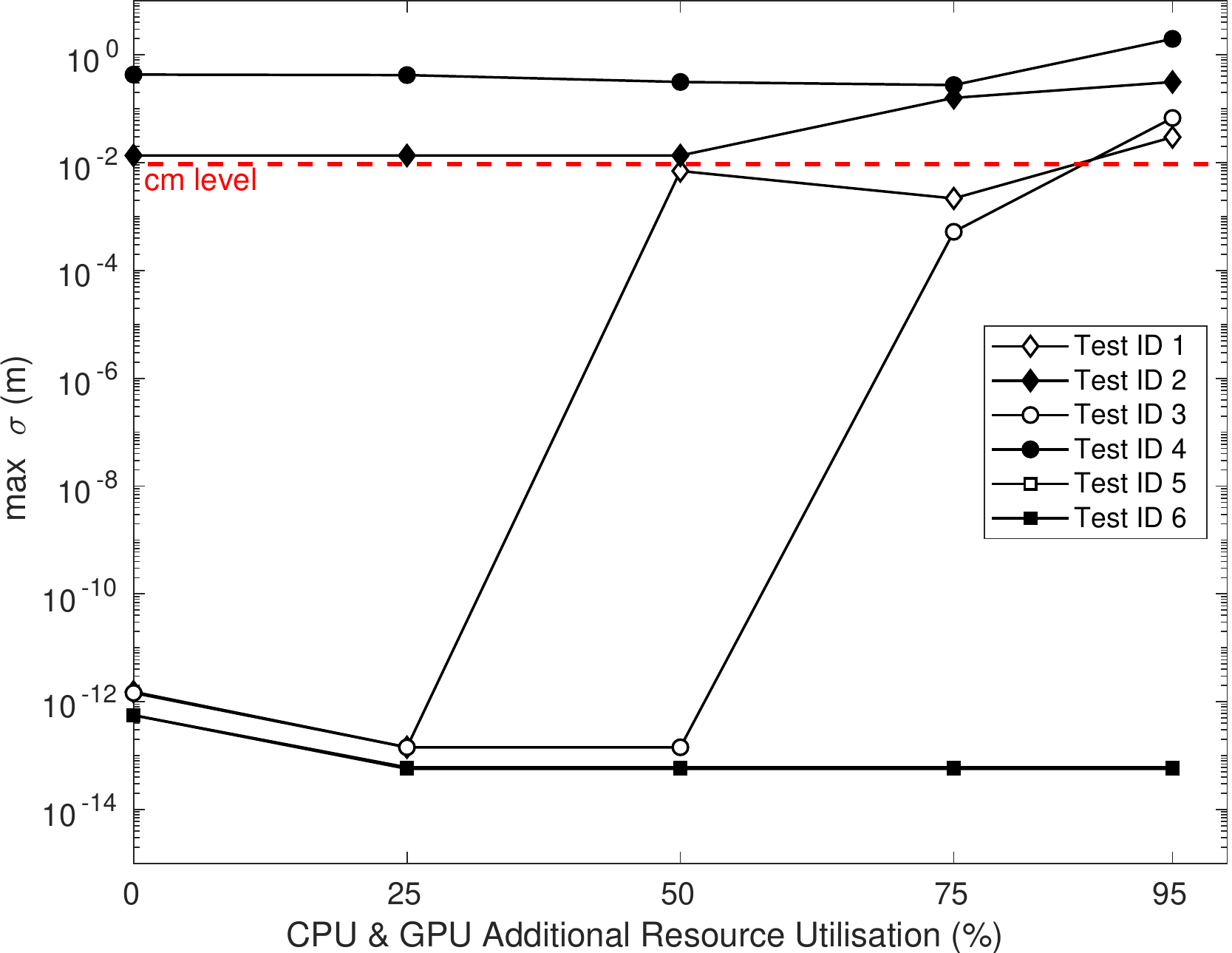}
    \caption{Summary of results showing maximum deviation for each scenario against different resource utilisation levels. Tests 5 and 6 overlap having almost identical results. Note that the lines between data points are only a guide.}
    \label{ExperimentsStressSummary}
\end{figure}

% ***************************************************
%  System Configuration and Screening
% ***************************************************

\subsection{Experimental System Configuration and Pre-Screening}\label{s:screening}
The experiments were carried out on an Alienware Area 51 R5 with an i9 9960X processor with 64GB non-ECC DDR4 RAM at 2933MHz with an NVIDIA GeForce RTX 2080 GPU with 8GB GDDR6 RAM at 1515 MHz. The operating systems was Linux Ubuntu 18.04.4 LTS. Tests were carried out in CARLA (v0.9.6 and Unreal Engine v4.22) using synchronous mode with a fixed $dt$ of $0.05$s. 

Initial testing~\cite{TSLUnrealEngineTesting} indicated an actor path deviation of $1\times10^{-13}$cm for 997 out of 1000 tests, with three tests reporting a deviation of over {\raise.17ex\hbox{$\scriptstyle\sim$}}$10$cm. While executing 100 repeats may seem sufficient, this sample size may fail to observe events that occur with low probability, giving false confidence in the results. Therefore each test was repeated 1000 times to provide a sufficient sample size. 
Due to the low probability of a simulation trace exceeding the permissible variance, using an average would `hide' these errors; hence this is the reason for using maximum variance and not an average.
A detailed guide for reproducing the experiments along with the scripts used are provided on github\footnote{\url{https://github.com/TSL-UOB/CAV-Determinism/tree/master/CARLA_Tests_setup_guide}}. To eliminate some of the potential sources of non-determinism outlined in Section~\ref{s:nondeterminisimSources} a series of screening tests and analyses were performed on our system. These were:

\begin{itemize}[leftmargin=*]
    \item System memory: \texttt{memtest86}~\cite{MemTest86} full suite of tests ran, all passed.
    \item Graphical memory: \texttt{cuda\_memtest}~\cite{cuda_memtest}, no failures on all 11 tests~\cite{shi2009testing}.
\end{itemize}

\begin{figure*}[t]
    \centering
    \begin{subfigure}{.49\textwidth}
        \includegraphics[width=1\textwidth]{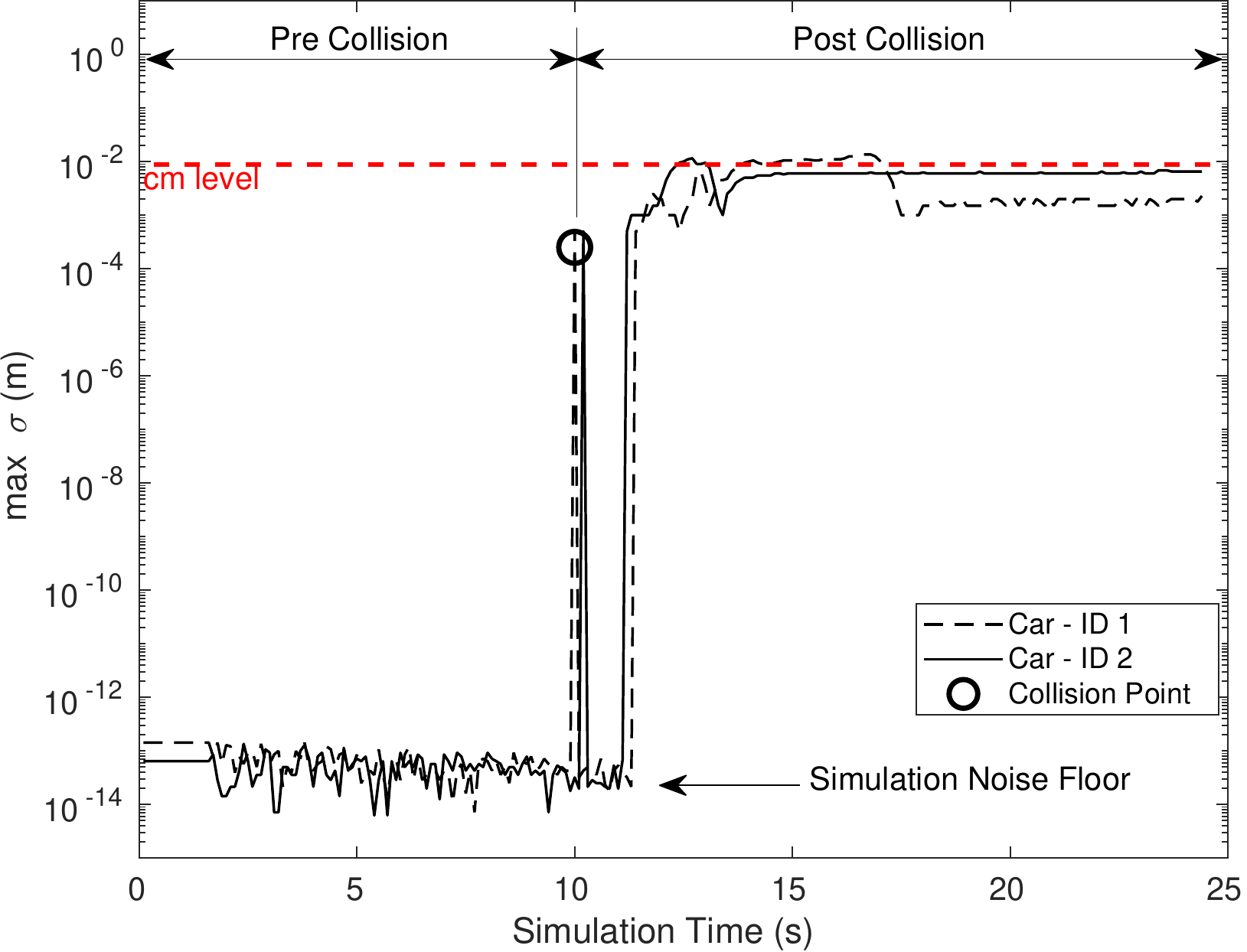}
        \caption{}
        \label{CarsCollisionCG25}
    \end{subfigure}
    \begin{subfigure}{.49\textwidth}
        \includegraphics[width=1\textwidth]{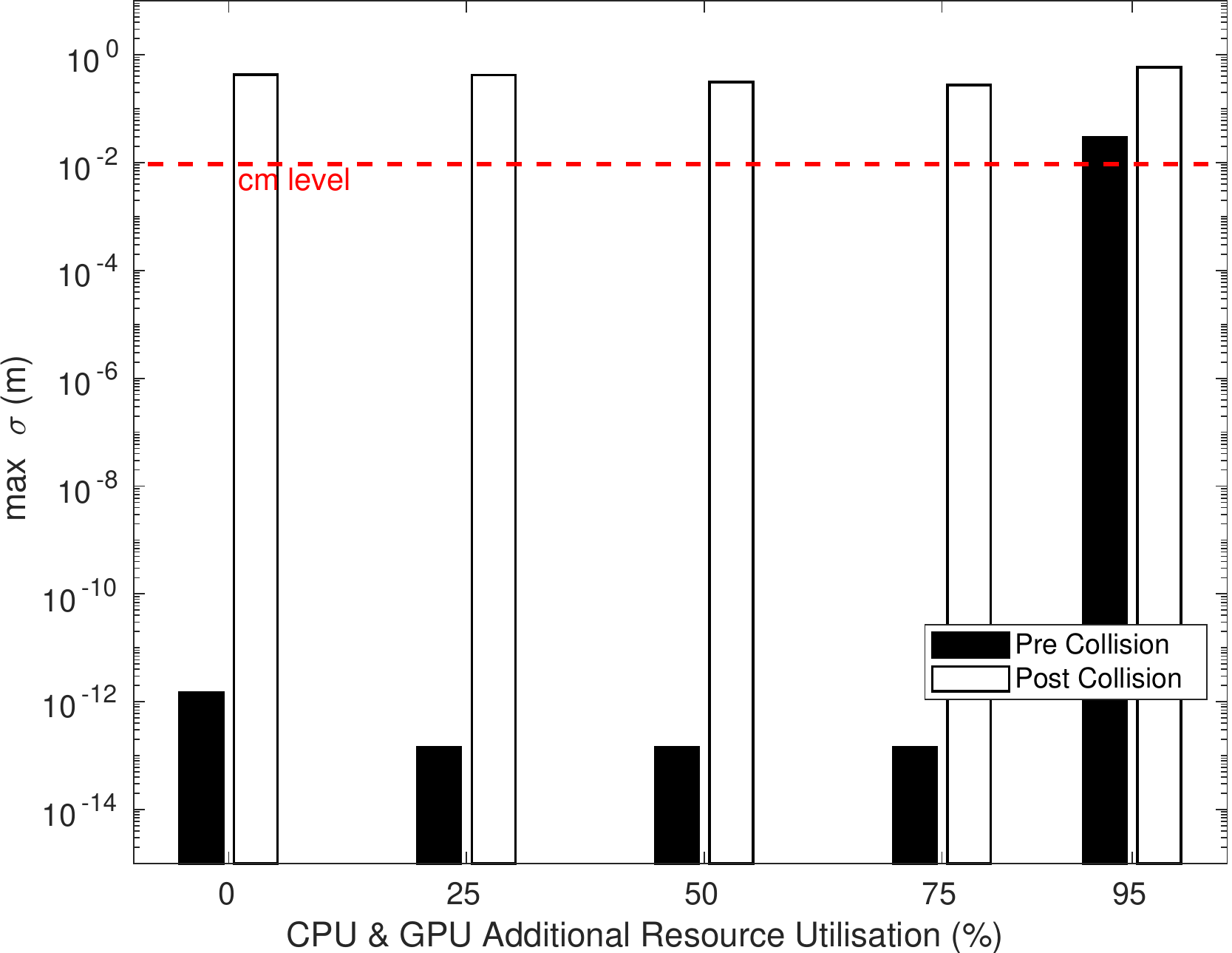}
        \caption{}
        \label{CarsCollisionPrePost}
    \end{subfigure}
    \caption{Vehicle to vehicle collision (Test 2) showing (a) maximum deviation against simulation time for 25\% resource utilisation and (b) maximum deviation pre- and post-collision against resource utilisation. The simulation noise floor is shown in (a) which is the empirical lower limit of deviation for the hardware reported in this study.}
\end{figure*}

\begin{figure*}[h]
    \centering
    \begin{subfigure}{.49\textwidth}
        \includegraphics[width=1\textwidth]{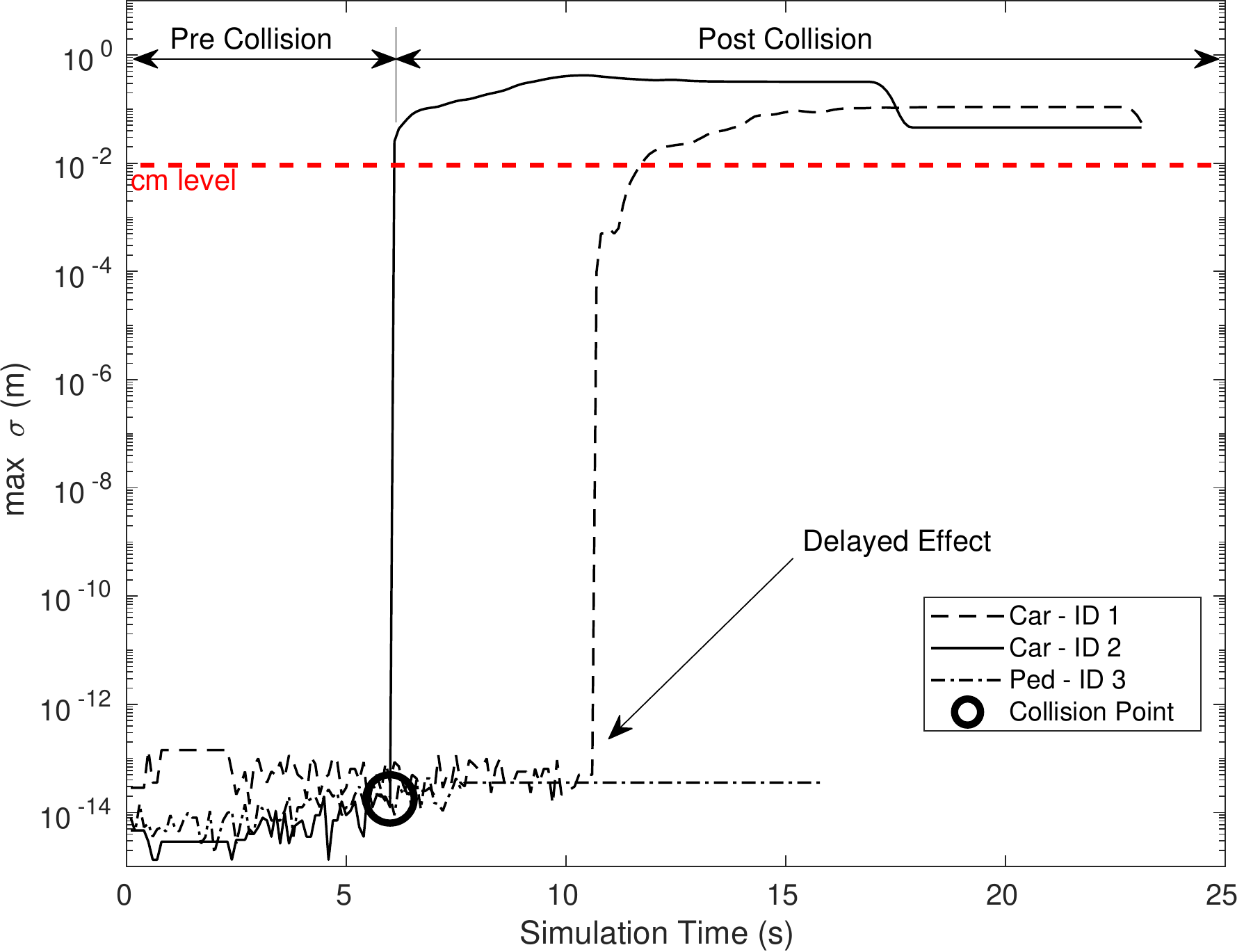}
        \caption{}
        \label{CarsPeopleCollsionsCG25}
    \end{subfigure}
    \begin{subfigure}{.49\textwidth}
        \includegraphics[width=1\textwidth]{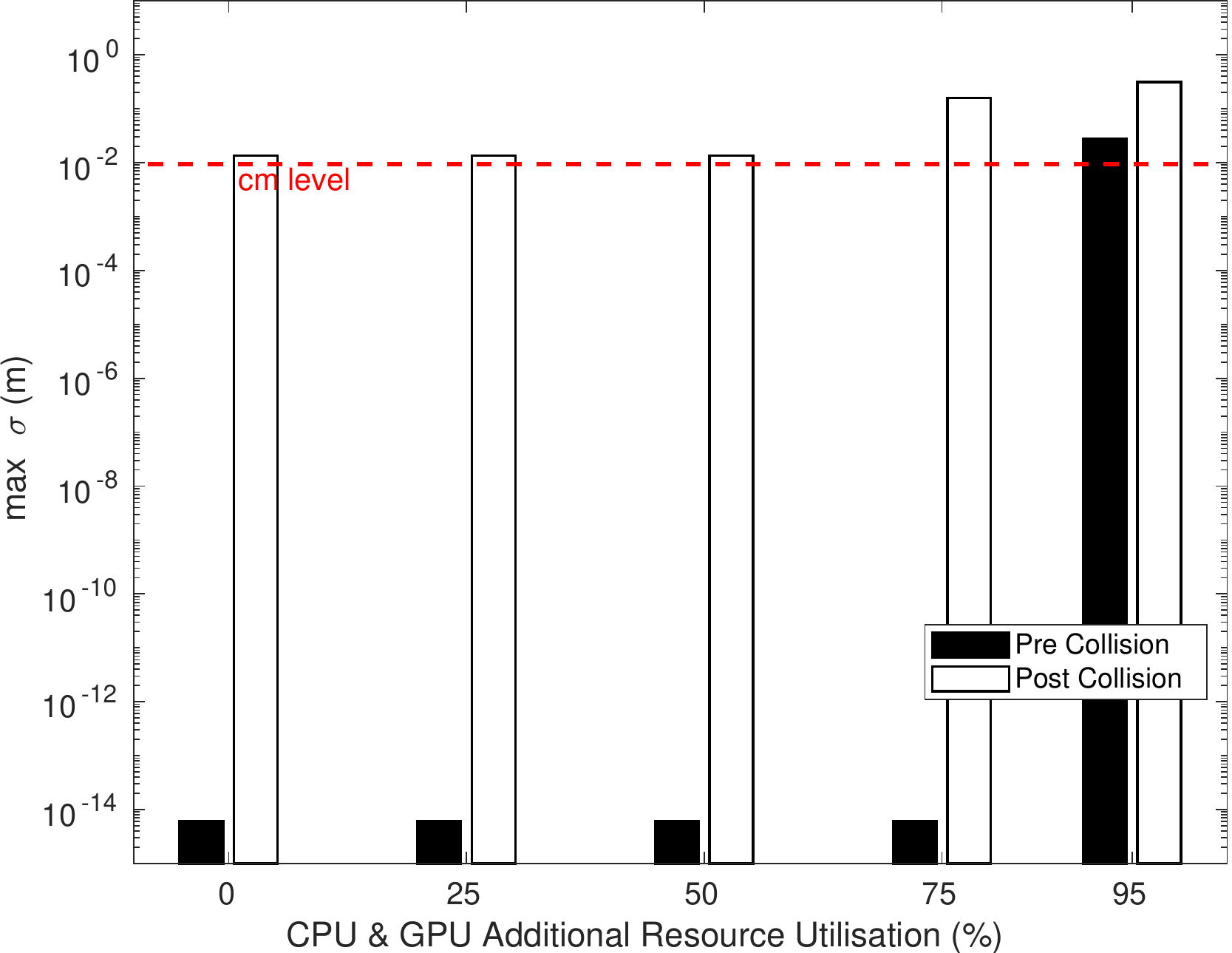}
        \caption{}
        \label{CarsPeopleCollisionPrePost}
    \end{subfigure}
    \caption{Vehicle to pedestrian collision (Test 4) showing (a) maximum deviation against simulation time for 25\% resource utilisation and (b) maximum deviation pre- and post-collision for different resource utilisation levels.}
\end{figure*}

A summary of the main results is shown in Table~\ref{TableOfExperiments}.  In the column $\max\sigma$ (unrestricted) the value reported is the maximum deviation across all resource utilisation levels, i.e.\ the worst case for a given scenario. From these results it is clear that scenarios with only pedestrian actors (Tests 5 \& 6) display results within tolerance over all resource utilisation levels with or without a collision where $\max\sigma$ is $5.6\times10^{-13}$ or $0.56\si{\pico\metre}$. However, all other scenarios involving vehicles or a mixture of actor types do not meet the required tolerance, with some deviation in actor path as large as $59$cm. 
A plot of actor position in the X--Y plane (plan view, units $m$) is shown in Fig.\ref{actorPathPlot}, where the inset clearly shows the divergence of the path of vehicle 2 post-collision with the pedestrian.
Clearly, such a large deviation cannot be acceptable for simulation to be considered a credible verification tool.

Resource utilisation was found to have a significant impact on \textit{simulation variance}. Figure~\ref{ExperimentsStressSummary} shows $\max\sigma$ against the artificially increased resource utilisation level, where the $x$-axis indicates the approximate percentage of resource utilisation (for CPU \& GPU). In this figure, any $\max\sigma$ above the $1$cm level (indicated by a dashed line) is considered non-permissible according to our specified tolerance. Note that the non-permissible results in Figure~\ref{ExperimentsStressSummary} (all those above the dashed line) are the worst case account of the situation, as per Equation~\ref{eq:max_sigma}, as the maximum variance is taken over the entire simulation period.

A general pattern in the results indicates that some scenarios consistently fail to produce results within tolerance, irrespective of resource utilisation (cf.\ Fig.~\ref{ExperimentsStressSummary} Test 2 \& 4 are above the dashed $1$cm line), while some are consistently within tolerance (cf.\ Fig.~\ref{ExperimentsStressSummary} Test 5 \& 6 both are with pedestrians only), and some cases only fail to meet the tolerance requirement at higher resource utilisation levels, i.e.\ above 75\% resource utilisation (cf.\ Fig.~\ref{ExperimentsStressSummary} Test 1 \& 3). 

Examining specifically the results from Tests 2 \& 4 as a function of simulation time reveals further information about the simulation variance before and after an actor collision. Fig.~\ref{CarsCollisionCG25} shows this examination for vehicle to vehicle collisions (Test 2), where $\max\sigma$ switches from permissible prior to the vehicle collision to non-permissible post collision. 
The pattern of permissible results prior to collision and non-permissible post collision is maintained up to a resource utilisation level of approximately 75\%, see Fig.~\ref{CarsCollisionPrePost}. This time series examination was repeated for vehicle to pedestrian collisions (Test 4) and the results are shown in Fig.~\ref{CarsPeopleCollsionsCG25}. Similarly to vehicle-to-vehicle collisions, the variation of $\max\sigma$ for vehicle to pedestrian collisions indicates permissible pre-collision behaviour with up to 75\% resource utilisation, see Fig.~\ref{CarsPeopleCollisionPrePost}. This is a key finding; it suggests that verification engineers should consider terminating tests at the point of a collision, as any post-collision results will be non-permissible.

The second key finding of this work is illustrated in Fig.~\ref{CarsPeopleCollsionsCG25}. In this scenario (Test 4), there is a collision between a vehicle (Car ID 2, solid line) and a pedestrian (Ped ID 3, dot dash line) which occurs at a simulation time of approximately $6$s and a second vehicle actor (Car ID 1, dashed line), which is \textit{not involved in the collision}. There are three observations; firstly that the vehicle directly involved in the collision (Car ID 2) displays high simulation variance immediately after the collision. Secondly, that the maximum deviation of the pedestrian involved in the collision (Ped ID 3) is at a tolerable level throughout the test\footnote{However, please note that in CARLA the pedestrian object is destroyed post-collision hence the flat line from $t=6$s onwards.}. Thirdly, we observed a delayed effect on Car ID 1 showing high simulation variance with a $5$s delay \textit{even though this vehicle was not involved in the collision}. This final point should be of particular concern to verification engineers, developers and researchers in the field as it implies that \textit{any collision between actors can affect the simulation variance of the entire actor population} and could potentially result in  erroneous simulation results. % not just assertion checking 

\begin{figure}[t]
    \centering
    \includegraphics[width=0.48\textwidth]{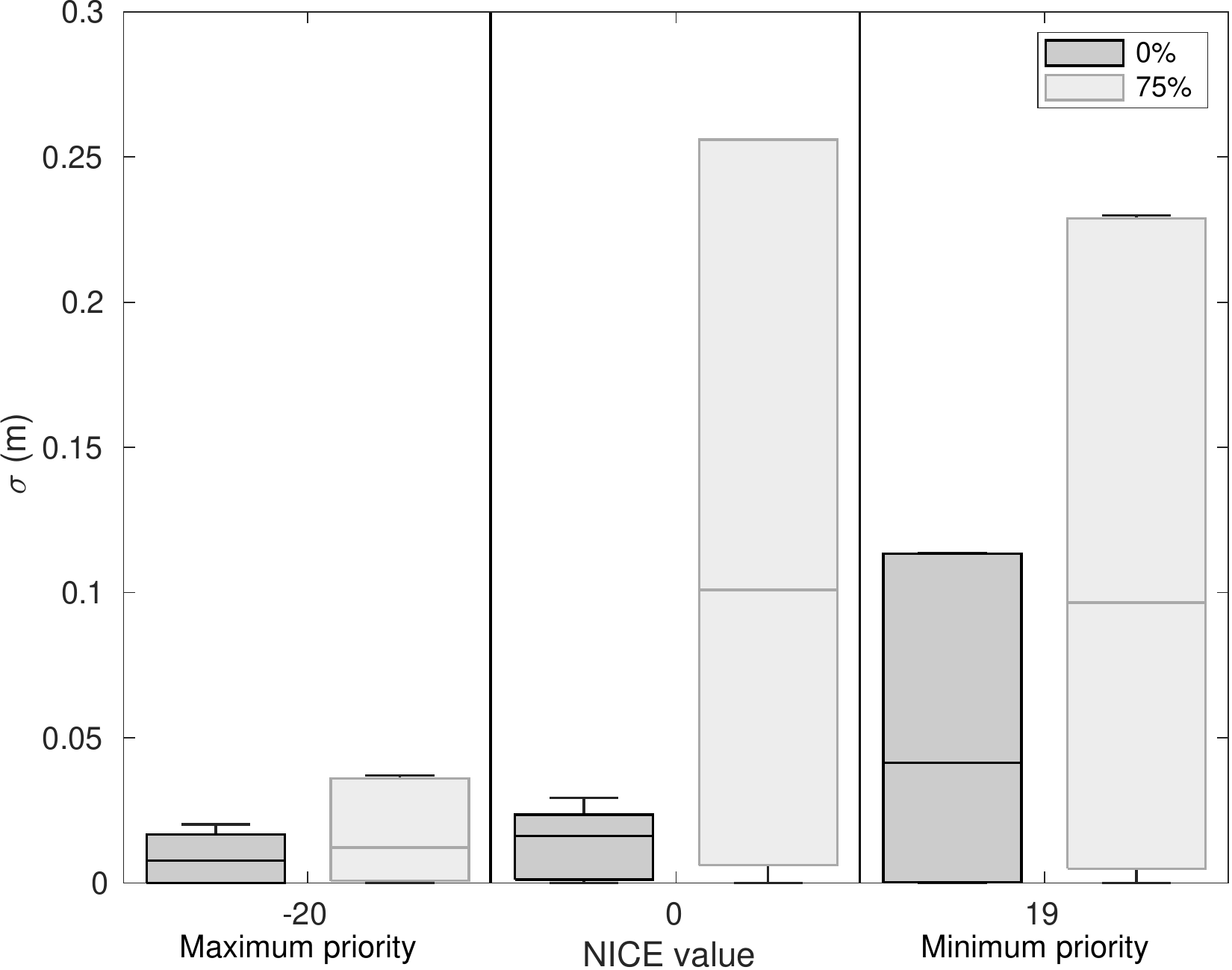}
    \caption{Variance range of three NICE priority settings for additional CPU \& GPU resource utilisation of 0\% and 75\%.}
    \label{NICEExperimentStressSummary}
\end{figure}

\begin{figure*}[t]
    \centering
    \includegraphics[width=0.99\linewidth]{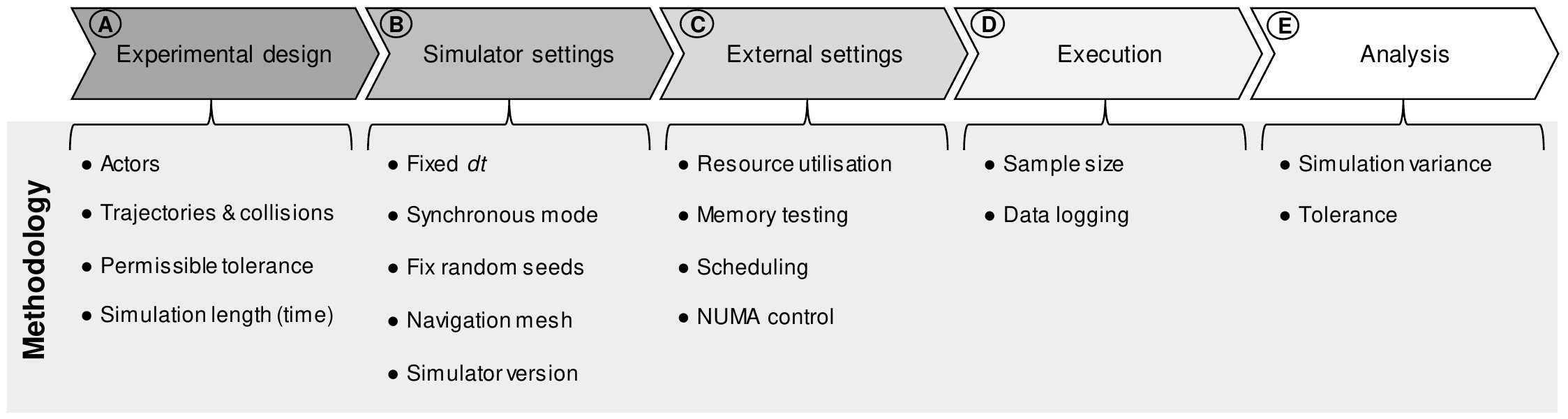}
    \caption{Stages of the method proposed to determine the variance of a simulation.}
    \label{method_diagram}
\end{figure*}

To conclude, the main findings of this case study suggest a working practice that would minimise the factors that give rise to the non-deterministic effects observed in this investigation. By limiting simulation results to pre-collision data and ensuring resource utilisation levels do not exceed 75\%, the permissible variance of $1$cm is achievable as shown in the \textit{restricted} column in Table~\ref{TableOfExperiments}. By applying this set of restrictions upon the simulation the maximum observed deviation across all experiments was $0.98$cm which is within the target tolerance we set out to achieve. 
Practitioners may wish to set a stricter resource utilisation level, such as less than 50\% to further reduce the potential dispersion of results if this is required for their chosen application. 

A correlation between resource utilisation and simulation variance has been observed in these results which may be due to process scheduling. Rather than limiting utilisation of the whole system, another approach may be to promote the scheduling priority of the simulation process which is explored in Section~\ref{r:process_scheduling}. Furthermore, we were keen to investigate the impact of memory access and include a brief investigation of this in Section~\ref{r:numa}.

% ***************************************************
%  Process Scheduling Priority
% ***************************************************

\subsection{Process Scheduling Priority} \label{r:process_scheduling}

An investigation into the impact of process scheduling on the simulation variance was undertaken following the observations of increased simulation variance with increased resource utilisation. The experiment was repeated ($n=1000$) using Test 1 but altering the process scheduling priority using the program \texttt{NICE}.\footnote{\url{http://manpages.ubuntu.com/manpages/bionic/man1/nice.1.html}} Setting a higher priority for the simulator process with respect to the resource utilisation processes, it was possible to determine if scheduling could account for the increased simulation variance when the system is under high resource utilisation.
To give a process a high priority a negative NICE value is set with the lowest being -20. To decrease the priority a positive value is set, up to +19. The default NICE value is 0.

The results are presented in Fig.~\ref{NICEExperimentStressSummary} where the box denotes the inter-quartile range of deviation, non-outlier limits by the whiskers and a horizontal bar for the median. The figure shows that decreasing the priority of the simulator process (right hand side of the  plot) has little effect on simulation variance when compared to a default NICE value of 0 (central bars in the plot). Increasing priority (left hand side of the plot) significantly reduced the variance for the 75\% resource utilisation experiment, but this does not account for all the difference in the observed results. This can be seen in the maximum priority setting where the bars in the plot are not equal, indicating an additional contribution to variance not accounted for by the NICE scheduling. 
The remaining difference in the variance between the two resource utilisation levels may be due to the lack of absolute control that \texttt{NICE} has over process scheduling.\footnote{\url{https://askubuntu.com/questions/656771/process-niceness-vs-priority}}

% ***************************************************
%  Non-uniform memory access (NUMA)
% ***************************************************

\subsection{Non-uniform memory access (NUMA)} \label{r:numa}
An additional investigation into memory access was undertaken given the potential impact that data transfer cost and execution order may have on simulation variance, see Section~\ref{s:sources_numa}. %
The program \texttt{numactl}~\cite{numactl_NUMA} allows a process to run with a specified memory placement policy, essentially allowing the process to be bound to a particular CPU core. %
\texttt{numactl} was used to fix the simulator and test script to single cores, and a (2\%) improvement in simulation variance was observed. %
This was considered minor comparative to the changes in simulation variance observed in other aspects of the case study, for example the pre- and post-collision between a vehicle and pedestrian saw a change of $10^{14}$. Therefore \texttt{numactl} was not used for subsequent testing.

However, practitioners that are aiming to minimise simulation variance should use a technique such as this to minimise data transfer cost and potentially minimise any potential effect on process execution order that may lead to increased simulation variance.

% ***************************************************
%  Investigation Summary
% ***************************************************

\subsection{Investigation Summary} \label{s:empirical_summary} 
These empirical investigations have highlighted the shortcomings of using a games engine for simulation based verification and suggests advice for best working practice. It was observed that resource utilisation positively correlates with simulation variance and that specific simulation events, such as vehicle collisions, can also lead to a breach in permissible tolerance. The investigation found that the effect of higher simulation variance as a result of increased resource utilisation can be reduced, but not omitted entirely, by controlling the scheduling policy. The investigation into specifying memory placement on the simulation process did improve simulation variance but only by a minor amount of 2\%.

However, these results are specific to the hardware and software used in the study and may not be transferable to other systems directly. Therefore we have derived a general methodology that practitioners can follow to find the \textit{operational domains of permissible variance} for a game-engine-based simulation environment. This methodology is presented in the next section.

% ***************************************************
%  METHOD
% ***************************************************

\section{Method to determine the Variance of a Simulation} \label{s:methodology}
In this section a method for determining the simulation variance of actor paths and resolving the operational domains of permissible variance is presented here as a work flow, see Fig.~\ref{method_diagram}. In addition, recommendations and best practice guidelines for minimising simulation variance are suggested.

The method consists of a sequence of five stages; experimental design, simulator settings, external settings, execution and analysis. In the following, each stage is described in detail with reference to the items listed for each stage in Fig.~\ref{method_diagram}.

\subsection{Experimental Design}\label{s:design_experiment}

The experiment design is based around a series of carefully selected tests that varies one of the sources of non-determinism whilst keeping all other parameters constant. Each test is repeated $n$ times and the analysis of the results provide a confidence with a degree of statistical certainty. By varying a single parameter and controlling all others, the simulation variance associated with each source of non-determinism can be found and addressed. 

\subsubsection{Actors} \label{s:actors}

All actors that could be included in the simulation should be tested, including any non-standard CARLA or bespoke actors including the ego vehicle, see Section~\ref{ego_actor}. 

\subsubsection{Trajectories and Collisions} 
Actor paths, or the sequence of actions required to generate paths, should be hard-coded to ensure repeatability. These paths should include collisions between actors and potentially collisions between actors and static scenery if this is likely to occur in the simulation or as part of the verification process. Actor paths without collision are also important to include as these will serve as a baseline to the other tests; to see if deviations between runs in actor paths increases with collisions. 

\subsubsection{Permissible Variance} \label{s:threshold}
The verification engineer should set the \textit{permissible variance}, also termed \textit{tolerance}.
The tolerance depends on the objectives of the simulation and the granularity at which the simulation environment operates. For example, this tolerance must be sufficiently small to enable accurate assertion checking and coverage collection, but not so small for assertion results to differ for repeated runs. In Section~\ref{s:tolerance} a tolerance of $1$cm was considered sufficient for urban scenario assertion checking. In practice, it may be necessary to determine this tolerance experimentally. 

\subsubsection{Simulation Time} 
The simulation time will depend on the actor paths and terminating conditions. The simulation time should be sufficient to record interactions between actors but not so long that the testing takes an inconvenient time to complete. In our empirical investigation a simulation time of $10-20$s was sufficient to monitor the distinct change in events such as the pre- and post-collision including the delayed effect seen by other actors, shown in Fig.~\ref{CarsPeopleCollsionsCG25} in Section~\ref{s:FinalResultsSection}. The termination conditions of the simulation can be set by, for example, actors reaching their final trajectory waypoints. 

\subsection{Simulator Settings}
The settings internal to the game engine or other simulation environment should be set to ensure a fixed physics time step, $dt$. If using CARLA, a small fixed value, say $0.05s$, can be set by using \texttt{setting.fixed\_delta\_seconds = 0.05}. In Unity, the default fixed time step is set to $0.02s$~\cite{MonoBehaviour_unity}. 

In CARLA, synchronous mode must be used to allow communication to external controllers which ensures no sensor data are passed out of order to the simulator which is particularly important if a complex ego controller is used~\cite{carla_sim_config}. 

The use of random numbers must be controlled through fixed seeds, resulting in pseudo-randomness. Random numbers might be used in the simulation environment to control variations of background effects, e.g.\ weather patterns, or the navigation of random pedestrian actors, external vehicle controllers or other clients connected to the simulation environment. Actors that navigate through the environment should use a fixed navigation mesh. The version number of the CARLA and Unreal environment has also be shown to affect results, see~\cite{TSLUnrealEngineTesting}. Therefore, ensuring a consistent version number throughout testing is also important.

\subsection{External Settings} \label{ego_actor}

\subsubsection{Resource Utilisation}
The resources available to the simulator have been shown to have a significant effect on the variance of the path of simulated vehicles. Thus, it is important to understand at what level of resource utilisation the system running the simulation becomes susceptible to \textit{simulation variance}.

CPU utilisation software, such as the linux workload generator \texttt{stress}, %which is a workload generator program, 
can be used to spawn workers on any number of cores or virtual threads on a system. This can be used to artificially increase the load on the system. For GPU utilisation, \texttt{gpu-burn} can be employed using the \texttt{fur test}. Different resolutions and multiple instances can be used to tune graphical utilisation levels~\cite{GPU_stress}. Reported values of resource utilisation can be obtained using the system monitors \texttt{htop} and \texttt{nvidia-smi} for CPU and GPU, respectively. These values should be added to the data logs. Alternatively, in place of artificial resource utilisation, multiple instances of the simulation could be executed simultaneously. However, the granularity of control with this approach may be reduced.

\subsubsection{Memory Testing}
Prior to experimental execution the system hardware should be tested for memory conformity and to ensure no single bit errors are occurring, see Section~\ref{s:screening}. For mainboard memory \texttt{memtest86}\ can be used on most platforms to run a series of pre-defined memory test patterns. This memory testing software can also be used for ECC enabled hardware. Similarly, to test memory on Nvidia based graphical adaptors \texttt{cuda\_memtest} can be used to ensure no memory errors exist.

\subsubsection{Scheduling}
We hypothesise that thread scheduling may be a major contributor to the non-deterministic results of the empirical study. However, gaining fine control over the scheduling policy and thread execution order is non-trivial~\cite{acm-q-rr-interview}. The operating system schedules threads according to a specified scheduling policy, potentially based on equal thread priority. Thus, in such a case, all tasks non-essential to the simulator should be terminated to prevent interference with the simulation. Assigning a higher priority to the simulator process may help to alleviate conflicting task scheduling which can be achieved by using, for example \texttt{TaskSettings.Priority} in Windows\footnote{\url{https://docs.microsoft.com/en-us/windows/win32/taskschd/tasksettings-priority}}
% ~\cite{TaskSettingWindows} 
or \texttt{NICE} in Linux\footnote{\url{https://linux.die.net/man/1/nice}}.
%~\cite{Nice_linux}.

\subsubsection{NUMA Control}
Control over a Non-Uniform Memory Access policy can be achieved using \texttt{numactl} for multi-core processors with shared memory. This control allows the simulator process to be fixed on a single core, reducing and unifying memory access time. Investigations in tests performed with NUMA control resulted in only minor improvements in simulation variance, see Section~\ref{s:screening}. Using a fixed memory placement policy may assist if simulation variance is borderline to the tolerance but only of the order of a few percent from our observations.

\subsubsection{Ego Vehicle Controller} 
An ego vehicle was not used in this study, but the impact of introducing this to the simulator can be considered here. The ego vehicle is seen as another actor in the simulation but care must be taken to ensure that any control algorithms, machine learning modules and processing pipelines are deterministic. We recommended that this be treated as a separate source of non-determinism and handled accordingly.

\subsection{Execution}

\subsubsection{Sample Size}
It is recommended that the chosen sample size, i.e.\ the number of repeated tests, is determined empirically. We recommend monitoring variance while increasing the sample size in orders of magnitude until there is confidence that $\max\sigma$ will not exceed the permissible variance for the verification process. Note that we suggest monitoring the maximum value of $\sigma$, not the average, because if even a single simulation run is outside of the permissible variance it would be critically important to detect this. Using an average may lead to false confidence in the verification result.

\subsubsection{Data Logging}
Unique identifiers should be assigned to each experiment, each repeat and each individual actor. 
The time-stamped actor positions should then be recorded at fixed time intervals throughout the simulation in order to determine the variance in actor path. Additional information should also be logged such as the CPU and GPU utilisation levels and engine specific metrics such as game loop latency.

\subsection{Analysis}
For each experiment, the maximum value of actor path deviation over all time samples and actors, $\max\sigma$, should be analysed to identify which of the candidate sources of non-determinism require restriction or control to reach the domain of permissible variance within the simulation environment. 

% ***************************************************
%  CONCLUSION
% ***************************************************
\section{Conclusions \& Future Work}\label{s:conclusion}

Game engines offer simulation environments that are used for the development and verification of autonomous driving functions. Determinism of a simulator is required to achieve repeatability, which is essential to find and fix software bugs efficiently, and to ensure simulation results are trustworthy. If a simulator is non-deterministic then practitioners should at least be aware of, and know how to find, the operational domains where \textit{simulation variance} is tolerable. 

An investigation into the CARLA simulator revealed a significant simulation variance for repeated tests with the same initial conditions and event history, indicating non-determinism of the simulation. We then researched, identified and discussed potential sources for non-determinism in this context. 
%
% A systematic case study of the CARLA simulator uncovered the actual factors that contribute towards greater simulation variance, giving rise to non-deterministic execution. 
%
In particular, actor collisions and system-level resource utilisation were identified as key contributors to increased simulation variance when using CARLA and we recommend monitoring these during simulation. Alternatively some commercial driving simulators claim to be fully deterministic, for instance RFpro~\cite{rFpro2021}. 
%
% To meet the recommended simulation variance for urban autonomous vehicle verification demonstrated in our case study, we recommend being aware of actor collisions and resource utilisation levels during simulation execution. 
%
% Alternatively, there are commercially available driving simulators that claim to be fully deterministic, e.g.\ RFPro~\cite{rFpro2021}, 
These may be more suitable if using a game engine does not provide a simulation variance sufficient for the verification requirements.

A general method to assess the actor path variance of a game-engine-based simulation environment was then proposed. The method can be used to find the \textit{domains of permissible variance} of a simulation environment for a given system configuration. This can give AV developers and verification engineers increased confidence in the simulation results and reduce debug time. As future work, the method can be extended to other simulation platforms and to criteria other than the actor path, e.g.\ actor orientation and any status indicators that may be of interest, also including actions, sequences and timings that may be useful for verification purposes.

An ambitious avenue for future work is the development of a deterministic simulator for AV development and verification. This requires controlling all potential sources of non-determinism, including randomness and scheduling, very similar to the development of the record-and-reply debugger \texttt{rr}~\cite{RR_link}, originally developed to catch low-frequency non-deterministically failing tests at Mozilla~\cite{acm-q-rr-interview}.

% ***************************************************
%  Thanks
% ***************************************************
\section*{Acknowledgement}
This research has in part been funded by the ROBOPILOT and CAPRI projects. Both projects are part-funded by the Centre for Connected and Autonomous Vehicles (CCAV), delivered in partnership with Innovate UK under grant numbers 103703 (CAPRI) and 103288 (ROBOPILOT), respectively. This research was also supported in part by the “UKRI Trustworthy Autonomous Systems Node in Functionality” under grant number EP/V026518/1.Special thanks to David May, Roger Shepherd, CFMS, Auroch Digital and Fortelix for productive discussions and support.

\balance

% ***************************************************
%  Bib
% ***************************************************
\printbibliography

\end{document}